\documentclass[conference]{IEEEtran}
\usepackage{cite}
\usepackage{amsmath,amssymb,amsfonts,amsthm}
\usepackage{algorithmic}
\usepackage{graphicx}
\usepackage{textcomp}
\usepackage{xcolor}
\def\BibTeX{{\rm B\kern-.05em{\sc i\kern-.025em b}\kern-.08em
    T\kern-.1667em\lower.7ex\hbox{E}\kern-.125emX}}

\newcommand{\norm}[1]{\Bigl\lVert#1\Bigr\rVert}

\usepackage[hyphens]{url}
\usepackage[pagebackref=true]{hyperref}
\usepackage{cleveref}

\usepackage{enumitem}

\usepackage{siunitx}
\usepackage{import}

\begin{document}
\bstctlcite{IEEEexample:BSTcontrol}
\title{Semi-supervised domain adaptation with CycleGAN guided by a downstream task loss}

\author{\IEEEauthorblockN{Annika Mütze}
\IEEEauthorblockA{\textit{IZMD \& School of Mathematics}\\ \textit{and Natural Sciences} \\
\textit{University of Wuppertal}\\
\textit{Wuppertal, Germany} \\
muetze@uni-wuppertal.de}
\and
\IEEEauthorblockN{Matthias Rottmann}
\IEEEauthorblockA{\textit{IZMD \& School of Mathematics}\\ \textit{and Natural Sciences} \\
\textit{University of Wuppertal}\\
\textit{Wuppertal, Germany} \\
rottmann@uni-wuppertal.de}
\and
\IEEEauthorblockN{Hanno Gottschalk}
\IEEEauthorblockA{\textit{IZMD \& School of Mathematics}\\ \textit{and Natural Sciences} \\
\textit{University of Wuppertal}\\
\textit{Wuppertal, Germany} \\
hanno.gottschalk@uni-wuppertal.de}
}

\maketitle

\begin{abstract}
Domain adaptation is of huge interest as labeling is an expensive and error-prone task, especially when labels are needed on pixel-level like in semantic segmentation. Therefore, one would like to be able to train neural networks on synthetic domains, where data is abundant and labels are precise. However, these models often perform poorly on out-of-domain images. To mitigate the shift in the input, image-to-image approaches can be used. Nevertheless, standard image-to-image approaches that bridge the domain of deployment with the synthetic training domain do not focus on the downstream task but only on the visual inspection level.
We therefore propose a ``task aware'' version of a GAN in an image-to-image domain adaptation approach.
With the help of a small amount of labeled ground truth data, we guide the image-to-image translation to a more suitable input image for a semantic segmentation network trained on synthetic data (synthetic-domain expert).
The main contributions of this work are 1) a modular semi-supervised domain adaptation method for semantic segmentation by training a downstream task aware CycleGAN while refraining from adapting the synthetic semantic segmentation expert
2) the demonstration that the method is applicable to complex domain adaptation tasks
and 3) a less biased domain gap analysis by using from scratch trained networks.
We evaluate our method on a classification task as well as on semantic segmentation. Our experiments demonstrate that our method outperforms CycleGAN  -- a standard image-to-image approach -- by 7 percent points in accuracy in a classification task using only 70 (10\%) ground truth images. For semantic segmentation we can show an improvement of about 4 to 7 percent points in mean Intersection over union on the Cityscapes evaluation dataset with only 14 ground truth images during training.
\end{abstract}


%%%%%%%%%%%%%%%%%%%%%%%%%%%%%%%%%%%%%%%%%%%%%%%%%%%%%%%%%%%%%%%%%
% INTRODUCTION
%%%%%%%%%%%%%%%%%%%%%%%%%%%%%%%%%%%%%%%%%%%%%%%%%%%%%%%%%%%%%%%%%
\section{Introduction} \label{sec:intro}
For automatically understanding complex visual scenes from RGB images, semantic segmentation -- pixel-wise classification of the image -- is a common but challenging method. The state-of-the-art results are achieved by deep neural networks (DNNs)\cite{chen_learning_2019, yuan_object-contextual_2020, tao_hierarchical_2020, dosovitskiy_image_2020, liu_swin_2021}. These models need plenty of labeled images to generalize well on unseen scenes. However, a manual label process on pixel level detail is time and cost consuming and usually error-prone\cite{cordts_cityscapes_2016, rottmann_automated_2022}. To reduce the labeling cost, methods which make use of weak or coarse labels like bounding boxes for the semantic segmentation task were proposed\cite{dai_boxsup_2015, Khoreva_simple_2017}.
Other approaches like semi-supervised learning\cite{van_engelen_survey_2020} reduce the amount of labeled ground truth needed. Thereby a large amount of unlabeled data is available but only a small subset has detailed labels\cite{hung_adversarial_2018, mondal_revisiting_2019, Ouali_semi-supervised_2020 }. With active learning one tries to query only the most informative images from a pool of unlabeled data to label and therefore reduce the label amount\cite{colling_metabox_2020, cai_revisiting_2021}. The combination of both ideas was studied by Rottmann et al.\cite{rottmann_deep_2018} for classification tasks.
However, these methods are limited to scenarios captured in the real world. Furthermore, the annotation cost of weak labels still might be intractable\cite{tsai_learning_2018}.
On the other hand, in the recent years simulations of urban street scenes were significantly improved\cite{richter_playing_2016, dosovitskiy_carla_2017, wrenninge_synscapes_2018}. The advantage of synthetic data is that images generated by a computer simulation often come with labels for the semantic content for free. To enlarge the real-world labeled dataset, synthetic data can be incorporated\cite{wrenninge_synscapes_2018}. Training on synthetic data has the potential to build a well performing network as plenty of data is available and diverse scenarios can be generated which are rare or life-threatening in the real world. However, switching the domain confuses the network\cite{hoffman_fcns_2016}. Even if the model learns to generalize well on one domain (e.g. real world) it can fail completely on a different domain (e.g. synthetic/images from simulation)\cite{wrenninge_synscapes_2018} or vice versa.
Domain adaptation (DA) is used to mitigate the so-called domain shift between one domain and another\cite{csurka_domain_2017}.
It aims at improving the model’s performance on a target domain by transferring knowledge
learned from a labeled source domain.
DA has become an active area of research in the context of
deep learning\cite{toldo_unsupervised_2020} ranging from adaptation on feature level\cite{tsai_learning_2018, vu_advent_2019,tsai_domain_2019}, adaptation on input level\cite{hoffman_fcns_2016, hoffman_cycada_2018, murez_image_2018, dundar_domain_2018, kang_pixel-level_2020, brehm_semantically_2022}, self-training\cite{mei_instance_2020, yang_fda_2020, kim_learning_2020, zhang_prototypical_2021}, a combination thereof\cite{kim_learning_2020} to semi-supervised approaches\cite{chen_semi-supervised_2021}.
Depending on the amount of labels available in the target domain the DA is unsupervised (UDA; no labels available), semi-supervised (SSDA; a few labels available) or supervised (SDA; labels exist for all training samples)\cite{toldo_unsupervised_2020}.
Adapting on input level to the style of the target domain disregarding the downstream task at hand but preserving the overall scene was firstly motivated by Hertzmann et al.\cite{hertzmann_image_2001} and is referred to image-to-image translation (I2I)\cite{zhu_unpaired_2017}.

Taking advantage of the synthetic domain we train a downstream task expert therein. We then shift the out-of-domain input (real world) closer to the synthetic domain via a semi-supervised image-to-image approach for mitigating the domain gap to the more abstract domain (e.g. simulation).  We thereby refrain from changing the expert which leads to a modular DA method.
Based on CycleGAN\cite{zhu_unpaired_2017}, we combine this GAN-based image-to-image method with a downstream task awareness in a second stage to adapt to the needs of the downstream task network. With the help of a very small contingent of ground truth (GT) in the real domain, we introduce the downstream task awareness of the generator.

Our main contributions:
\begin{itemize}
    \item we present a novel modular semi-supervised domain adaptation method for semantic segmentation under domain shift by guiding the generator of an image-to-image domain adaptation approach to a semantic segmentation task awareness. Thereby we refrain completely from retraining our downstream task network.
    \item we demonstrate that our method is applicable to multiple complex domain adaptation tasks.
    \item we consider a pure domain separation in our analysis by using from-scratch-trained neural networks leading to a less biased domain gap.
\end{itemize}
Based on our knowledge this is the first time the generator of a GAN setup is guided with the help of a semantic segmentation network to focus on the downstream task.

Furthermore, the modular composition of generator and semantic segmentation network can be understood as a method to establish an abstract intermediate representation in a data-driven manner. We study how well the generator can adapt to its tasks of generating the abstract representation and supporting the downstream task. The abstract representation of real world data thereby has the potential to lead to more powerful DNNs.

The remainder of this paper is organized as follows. In \cref{sec:relwork} we review existing approaches in the context of domain adaptation, particularly semi-supervised domain adaptation and image-to-image approaches. It follows a detailed description of our method in \cref{sec:method}. We then evaluate our method on two different tasks and three different datasets in \cref{sec:experiments}, showing considerable improvements with only a small number of GT data. A classification task on a real-to-sketch domain shift demonstrates the capacity of the method in terms of domain shift complexity and our detailed experiments on semantic segmentation of urban street scenes illustrate the potential of the method on more complex downstream tasks. For evaluation metrics we use the well established mean Intersection over Union\cite{jaccard_distribution_1912, everingham_pascal_2015} for semantic segmentation and report accuracy for classification experiments. Finally, we conclude and discuss our results as well as giving an outlook to future work in \cref{sec:discussion}.

%%%%%%%%%%%%%%%%%%%%%%%%%%%%%%%%%%%%%%%%%%%%%%%%%%%%%%%%%%%%%%%%%
% RELATED WORK
%%%%%%%%%%%%%%%%%%%%%%%%%%%%%%%%%%%%%%%%%%%%%%%%%%%%%%%%%%%%%%%%%
\section{Related Work} \label{sec:relwork}
Our work bases on two main concepts: domain adaptation via image-to-image translation and semi-supervised learning in the context of GANs and domain adaptation. In the following we will review the most recent literature on these topics.

Since its first motivation by Hertzmann et al.\cite{hertzmann_image_2001}, image-to-image translation is nowadays dominated by generative adversarial networks (GANs)\cite{goodfellow_generative_2014} conditioned on an image and has various applications in computer vision and computer graphics\cite{pang_image--image_2021}. The concept of GANs is to model the data generating distribution implicitly by training two adversary networks, a generator generating images and a discriminator predicting the input being generated or a 'real' training sample. Formerly, paired data was needed to adapt to the new style\cite{isola_image--image_2017}. But as paired data is hard to get or generate, unsupervised methods were developed and published like CycleGAN\cite{zhu_unpaired_2017}, where a cycle consistency loss leads to a consistent mapping of an image from one domain to the other.

Image-to-image in general can be evaluated by different methods\cite{pang_image--image_2021}, e.g., image structural similarity index (SSIM)\cite{wang_image_2004}, Fr\'{e}chet inception distance (FID)\cite{heusel_gans_2017} or LPIPS\cite{zhang_unreasonable_2018}. For I2I in the context of domain adaptation, as a fixed task is considered, the downstream task performance is used as performance measure for the image translation. Nevertheless, our generator output could also be evaluated with other metrics.

Our idea to use I2I for domain adaptation while generating more semantically consistent images with the help of the downstream task loss is also tackled, for example, in Cycada\cite{hoffman_cycada_2018}, SUIT\cite{li_simplified_2020}, the work of Guo et al.\cite{guo_gan-based_2020} and Brehm et al.\cite{brehm_semantically_2022} but in an unsupervised manner.

In contrast to unsupervised approaches, SSDA methods can make use of
a few labeled images in the target domain. There are several
SSDA approaches in the context of classification\cite{wu_dcan_2018, saito_semi-supervised_2019, kim_attract_2020, jiang_bidirectional_2020, mabu_semi-supervised_2021} and
Singh et al.\cite{singh_improving_2021} proposes an active learning-based method to determine which data should be labeled in the target domain.

SSDA for semantic segmentation tasks is considered less. For example, Wang et al.\cite{wang_alleviating_2020} propose to adapt simultaneously on a semantic and global level using adversarial training.
Chen et al.\cite{chen_semi-supervised_2021} propose a student teacher approach aligning the cross-domain features with the help of the intra-domain discrepancy of the target domain, firstly considered in the context of domain adaptation by\cite{kim_attract_2020}.

Training in a two stage manner, starting with an unsupervised training (also referred to `pre-training') to initialize good network parameters and then retrain in a second stage with a small amount of labeled data is a common semi-supervised learning technique\cite{van_engelen_survey_2020}. We transfer this concept to GANs of an image-to-image method.
In the context of general GANs, using pre-training is not new. Wang et al.\cite{wang_transferring_2018} for example, analyzed pre-training for Wasserstein GANs with gradient penalty\cite{gulrajani_improved_2017} in the context of image generation when limited data is available.
In the work of Zhao et al.\cite{zhao_leveraging_2020}, the usefulness of pre-training GANs is shown based on the GP-GAN\cite{mescheder_which_2018}.
Grigoryev et al.\cite{grigoryev_when_2022} gives an overview over when, why and which pre-trained GANs are useful. We follow their suggestion of pre-training both the generator and the discriminator but refrain from the suggestion of using ImageNet\cite{deng_imagenet_2009} pre-trained GANs to not distort the domain gap analysis.

For the downstream task model we also train completely from scratch for a pure domain separation and keep the task network fixed after it is trained once which makes it reusable when switching to a different domain. This is in contrast to the other above-mentioned UDA approaches which use ImageNet pre-trained downstream task networks and adapt the downstream task network to mitigate the domain gap.
In general, UDA methods lack important information of the target domain compared to their supervised trained counterparts \cite{chen_semi-supervised_2021}. Furthermore, pure image-to-image methods are task agnostic and therefore may lack semantic consistency\cite{toldo_unsupervised_2020,guo_gan-based_2020}.
For this reason we propose a SSDA method with a task aware image-to-image component. Independently of our approach, Mabu et al.\cite{mabu_semi-supervised_2021} published recently a similar approach based on classification in a medical context with domain gaps primarily in image intensity and contrast.
On the contrary, our method aims at more demanding tasks such as semantic segmentation on much broader domain gaps like realistic to abstract domains leading to potentially broader applications.
Furthermore, our method allows an analysis of the influence of the task awareness compared to the standard loss.
In addition, we consider image-to-image for real images to the synthetic domain to retain the benefits of a synthetic expert, whereas the other approaches consider the opposite direction.

In the following we explain our method in more detail.

%%%%%%%%%%%%%%%%%%%%%%%%%%%%%%%%%%%%%%%%%%%%%%%%%%%%%%%%%%%%%%%%%
% METHODOLOGY
%%%%%%%%%%%%%%%%%%%%%%%%%%%%%%%%%%%%%%%%%%%%%%%%%%%%%%%%%%%%%%%%%
\section{Methodology} \label{sec:method}
\noindent Our method consists of three stages which are depicted in
fig. 1 and explained in detail in this section.
\begin{figure}
    \centering
    \scriptsize
    \def\svgwidth{\columnwidth}
    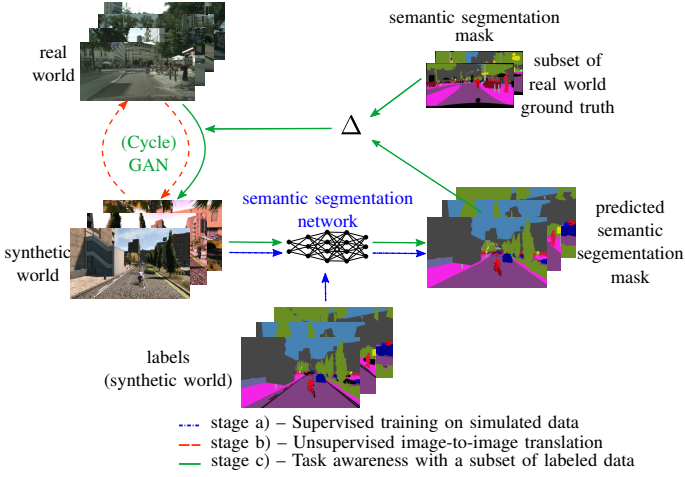
    \caption{Concept of our method: \textcolor[rgb]{0,0,1}{Stage a)} -- Training of a downstream task model (e.g. semantic segmentation network) on the abstract/synthetic domain.
    \textcolor[rgb]{1,0.19921875,0}{Stage b)} -- Training a CycleGAN based on unpaired data to transfer real data into the synthetic domain.
    \textcolor[rgb]{0,0.64453125,0.214843755}{Stage c)} -- We freeze the downstream task network and tune the generator with the help of a few labeled data points by guiding it based on the loss of the downstream task network.}
    \label{fig:concept}
\end{figure}

\paragraph{Training of downstream task network}
We assume that we have full and inexhaustible access to labeled data in the synthetic domain $\mathcal{S}$. Based on a labeled training set $\mathcal{T_{\mathcal{S}}} = \{(x^s_i, y^s_i) \in \mathcal{S} \times \mathcal{Y} : i=1, \ldots, N\}$, we train a neural network $f$ in a supervised manner on the synthetic domain solving the desired task (e.g., semantic segmentation or classification).
In contrast to the ``common practice''\cite{kang_pixel-level_2020}, we do not use ImageNet pre-trained weights for the downstream task network backbone. To ensure a pure domain separation and a non-biased downstream task network, we train $f$ completely from scratch.
This ensures that the network learns only based on the synthetic data, and we prevent a bias towards the real world. As a consequence, we accept a reduction of the total accuracy when evaluating the model on the real domain (out-domain accuracy). However, with the help of an independent validation set, we measure our in-domain accuracy to ensure appropriate performance in the synthetic domain.
After the model has reached the desired performance, we freeze all parameters and keep our synthetic domain expert model fixed.

\paragraph{Unsupervised Image-to-Image Translation}
To mitigate the domain gap between real and synthetic data, we build on the established image-to-image method CycleGAN\cite{zhu_unpaired_2017} -- a GAN approach which can deal with unpaired data by enforcing a cycle consistency between two generators, one for each domain ($G_{\mathcal{S}\to\mathcal{R}}$, $G_{\mathcal{R}\to\mathcal{S}}$). The domain discriminators to classify whether the sample is generated or an in-domain sample are denoted with $D_\mathcal{S}$ and $D_\mathcal{R}$. Let $x^r \sim p_{\text{data}}$ denote the data distribution in the real domain, then we can define the four loss components of the image-to-image translation of CycleGAN:
\begin{enumerate}
    \item \emph{adversarial loss for the generators}
        \begin{align}
             \mathcal{L}_{G_{\mathcal{R}\to\mathcal{S}}}& \nonumber \\
             = & \ \mathbb{E}_{x^r \sim p_{\text{data}}(x^r)} \left[ \Bigl(D_\mathcal{S}\Bigl(G_{\mathcal{R}\to\mathcal{S}}(x^r)\Bigr) -1 \Bigr)^2\right] \label{eq:genloss}
        \end{align}
        For $\mathcal{L}_{G_{\mathcal{S}\to\mathcal{R}}}$ substitute $\mathcal{R}$ resp. $r$ with $\mathcal{S}$ resp. $s$ in \cref{eq:genloss}. \\
    \item \emph{cycle consistency loss}:
        \begin{align}
            \mathcal{L}_{\text{cyc}}&(G_{\mathcal{R}\to\mathcal{S}}, G_{\mathcal{S}\to\mathcal{R}}) \nonumber\\
             = & \ \mathbb{E}_{x^r\sim p_{\text{data}}(x^r)}\left[\norm{G_{\mathcal{S}\to\mathcal{R}}\Bigl(G_{\mathcal{R}\to\mathcal{S}}(x^r)\Bigr)-x^r}_1\right]  \label{eq:cycleloss} \\
            + & \  \mathbb{E}_{x^s\sim p_{\text{data}}(x^s)}\left[\norm{G_{\mathcal{R}\to\mathcal{S}}\Bigl(G_{\mathcal{S}\to\mathcal{R}}(x^s)\Bigr)-x^s}_1\right].\nonumber
        \end{align}
    \item \emph{identity preserving loss}
        \begin{align}
            \mathcal{L}_{\text{identity}}&(G_{\mathcal{R}\to\mathcal{S}}, G_{\mathcal{S}\to\mathcal{R}}) \nonumber \\
            =& \ \mathbb{E}_{x^s\sim p_{\text{data}}(x^s)}\left[\norm{G_{\mathcal{R}\to\mathcal{S}}(x^s) - x^s}_1\right] \label{eq:idloss}\\
            + & \ \mathbb{E}_{x^r\sim p_{\text{data}}(x^r)}
            \left[\norm{G_{\mathcal{S}\to\mathcal{R}}(x^r) - x^r}_1\right]\nonumber
        \end{align}
    \item \emph{adversarial loss for the discriminators}
        \begin{align}
             \mathcal{L}_{D_\mathcal{S}} =  & \
             \mathbb{E}_{x^s \sim p_{\text{data}}(x^s)} \left[\Bigl(D_\mathcal{S}(x^s)-1\Bigr)^2\right] \nonumber\\
             + & \ \mathbb{E}_{x^r \sim p_{\text{data}}(x^r)}\left[D\Bigl(G_{\mathcal{R}\to \mathcal{S}}(x^r)\Bigr)^2\right] \label{eq:discrimloss}
        \end{align}
        For $\mathcal{L}_{D_{\mathcal{R}}}$ substitute $\mathcal{R}$ resp. $r$ with $\mathcal{S}$ resp. $s$ in \cref{eq:discrimloss}.
\end{enumerate}
\noindent
As adversarial losses $\mathcal{L}_{G_*}$ we use the least-squares loss proposed by Mao et al.\cite{mao_least_2017} which has been used in the implementation of CycleGAN and leads to a more stable training according to Zhu et al.\cite{zhu_unpaired_2017}.
With the help of the adversarial losses\cite{mao_least_2017} (\cref{eq:genloss} and \cref{eq:discrimloss}), the general aim of matching the target data distribution with the generated data is pursued. Furthermore, with the help of the cycle consistency loss the method enforces an approximate inverse mapping between the two domains.
In addition, the identity loss is used to prevent the generator to modify an image if it already lies within the desired domain.
The overall generator loss is defined as the weighted sum of the losses \cref{eq:genloss,eq:cycleloss,eq:idloss}:
\begin{align*}
    \mathcal{L}_{\text{Gen}} = \mathcal{L}_{G_{\mathcal{R}\to\mathcal{S}}}
    + \mathcal{L}_{G_{\mathcal{S}\to\mathcal{R}}}
    + \lambda_\text{cyc} \mathcal{L}_{\text{cyc}}
    + \lambda_\text{cyc}\lambda_{\text{id}} \mathcal{L}_{\text{identity}}
\end{align*}
This leads to a solid image-to-image translation. However, this translation is still task agnostic and therefore potentially misses important features when transferring the style from one domain to another.

\paragraph{Downstream task awareness}
We use the unsupervised models from stage b) as initialization for stage c) where we extend the model training and guide the generator with the help of a small amount of labeled data to the downstream task. Let $\mathcal{T_{\mathcal{R}}} = \{(x^r_i, y^r_i) \in \mathcal{R} \times \mathcal{Y} : i=1, \ldots, N_L\}$ be a labeled subset from domain $\mathcal{R}$, where $N_L$ denotes the number of labeled samples.

We achieve the downstream task awareness by extending the adversarial loss \cref{eq:genloss} for the generator $G_{\mathcal{R}\to\mathcal{S}}$ based on the loss of the downstream task network $f$. Given a problem with $C$ different classes and an input $x$, we calculate the loss between the prediction $f(x) = (f(x)_1, \ldots, f(x)_C)$ and the corresponding ground truth class label $y$ with the categorical cross entropy loss (also known as softmax loss):
\begin{align}\label{eq:taskloss}
    \mathcal{L}_\mathit{task}(x,y) &= \mathcal{L}_\mathit{CE} \Bigl(f(x),y\Bigr) \nonumber\\
    &= - \log\left( \dfrac{\exp({f(x)_y})}{\sum_{c=1}^C \exp(f(x)_c) }\right),
\end{align}
where $f(x)_y$ is the value assigned to the true class $y$ by the neural network. Depending on the task we additionally average the result. For example, in the case of semantic segmentation we calculate the mean over all pixel.
When using a batch size greater than $1$, the mean is additionally computed with respect to the batch size.

For a labeled training sample $t_i = (x^r_i, y^r_i) \in \mathcal{T_{\mathcal{R}}}$ we define the extended generator loss $\Tilde{\mathcal{L}}_{G_{\mathcal{R}\to\mathcal{S}}}$ with the help of a weighting factor $\alpha \in [0,1]$ as follows:
\begin{align}
    \Tilde{\mathcal{L}}_{G_{\mathcal{R}\to\mathcal{S}}} (t_i) = & (1 - \alpha) \: \underbrace{\Bigl(D_s\Bigl(G_{\mathcal{R}\to\mathcal{S}}(x^r_i)\Bigr) -1 \Bigr)^2}_{\substack{\text{adversarial loss as in \cref{eq:genloss}}}} \nonumber \\
     + &\:\alpha \: \underbrace{\Bigl( \mathcal{L}_\mathit{CE}\Bigl(f(G_{\mathcal{R}\to\mathcal{S}}(x^r_i)),y^r_i\Bigr) \Bigr)}_{\substack{\text{task loss}}}.
     \label{eq:extendedloss}
\end{align}
We use $\alpha$ for a linear interpolation between the two loss components to control the influence of one or the other loss during training.
The overall generator loss therefore becomes:
\begin{align*}
    \Tilde{\mathcal{L}}_{\text{Gen}} = \Tilde{\mathcal{L}}_{G_{\mathcal{R}\to\mathcal{S}}} + \mathcal{L}_{G_{\mathcal{S}\to\mathcal{R}}} + \lambda_\text{cyc} \mathcal{L}_{\text{cyc}} + \lambda_\text{cyc}\lambda_{\text{id}} \mathcal{L}_{\text{identity}}
\end{align*}
The discriminator losses (cf. \cref{eq:discrimloss}) are kept identically.

The combination of stage b) and c) leads to our semi-supervised learning strategy for the GAN training. For the DA the images generated by $G_{\mathcal{R}\to\mathcal{S}}$ are fed to $f$.
In principle our approach is independent of the chosen architecture as the general concept is transferable, and we make no restriction to the underlying domain expert as long as a task loss can be defined. Furthermore, due to our modular composition, the intermediate representation generated by $G_{\mathcal{R}\to\mathcal{S}}$ could also be evaluated with respect to other metrics described in\cite{pang_image--image_2021}.

%%%%%%%%%%%%%%%%%%%%%%%%%%%%%%%%%%%%%%%%%%%%%%%%%%%%%%%%%%%%%%%%%
% EXPERIMENTS
%%%%%%%%%%%%%%%%%%%%%%%%%%%%%%%%%%%%%%%%%%%%%%%%%%%%%%%%%%%%%%%%%

\section{Numerical Experiments} \label{sec:experiments}
We evaluate our method on two different downstream tasks: classification and semantic segmentation. For the first mentioned we consider the domain shift between real objects and their sketches and for the semantic segmentation we examine experiments on real world urban scenes transferred to two different simulations.

\subsection{Classification on real and sketch data}
\begin{figure}
    \centering
    \includegraphics[trim= 0 8 0 0, clip, width=0.45 \textwidth]{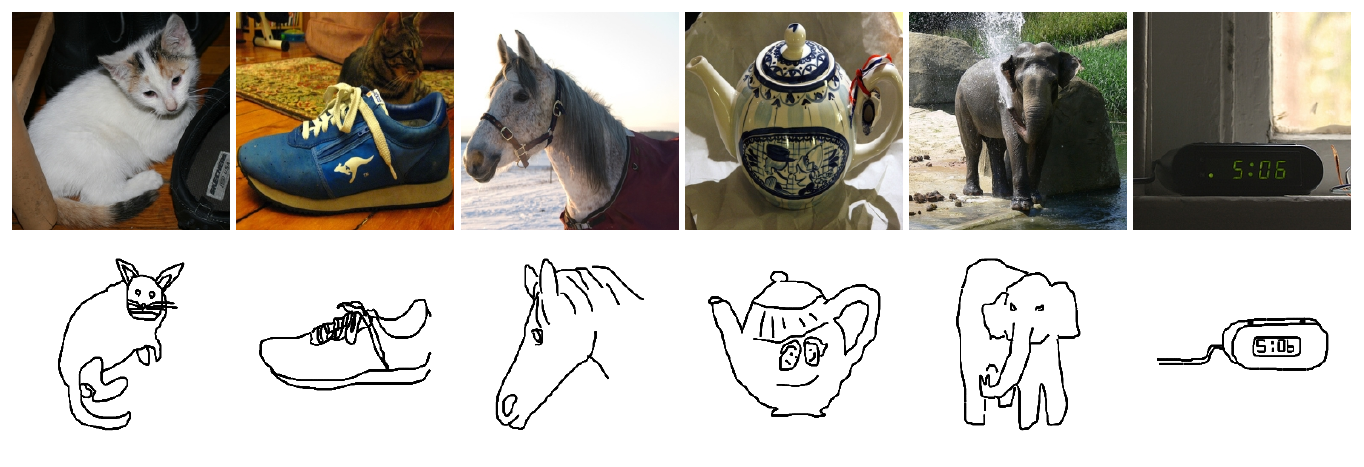}
    \caption{Examples of the Sketchy dataset. Top row: real photos. Bottom row: one of the corresponding sketches.}
    \label{fig:Sketchydataset}
\end{figure}

For the classification experiments we choose sketches as abstract representation of real world objects. Therefore, we consider a subset of the Sketchy dataset\cite{sangkloy_sketchy_2016}.
The original dataset comprises $125$ categories -- a subset of the ImageNet classes -- and consists of $12,\!500$ unique photographs of objects as well as $75,\!471$ sketches drawn by different humans. Per photograph, five participants were asks to draw a corresponding sketch to capture the diversity of sketching and drawing skills.
A detailed description of the dataset generation process is given in\cite{sangkloy_sketchy_2016}.
Fig.\ \ref{fig:Sketchydataset} shows examples from the dataset.
For our experiments we limit our dataset to the $10$ classes \textit{alarm clock}, \textit{apple}, \textit{cat}, \textit{chair}, \textit{cup}, \textit{elephant}, \textit{hedgehog}, \textit{horse}, \textit{shoe} and \textit{teapot}. As the original dataset includes sketches which are ``incorrect in some way''\cite{sangkloy_sketchy_2016}, we removed sketches which we could not identify as the labeled class. For validation, we randomly chose $50$ sketches per class for validation for the sketch data. As the number of real photos is more limited we chose $10$ random photos per class for validation. This results in a remaining training set of $4,\!633$ sketch images and $700$ real photos.

For the stage a) training we use a ResNet18\cite{he_deep_2016} as classifier which is our downstream task network. For the I2I based on CycleGAN (stage b) and c)) we used the implementation of\cite{zhu_unpaired_2017} and extended it according to our method described in \cref{sec:method}.
We fix the amount of GT data used in the stage c) training to $70$ images ($10\%$ of the data) for our experiment and use the categorical cross entropy loss as task loss.
We measure the performance of the classifier in terms of accuracy on a validation set $\mathcal{V} = \{(x_i, y_i) \in \mathcal{X} \times \mathcal{Y} : i=1, \ldots, N_v\}$ where $\mathcal{X}$ is a subset of $\mathcal{S}$ or $\mathcal{R}$.
\begin{align*}
    \mathit{acc}_f(\mathcal{V}) = \frac{\vert {\{(x,y) \in \mathcal{V} \mid f(x)=y\}}\vert}{\vert {\{(x,y) \in \mathcal{V}\}\vert}}.
\end{align*}

After stage a) training on sketch images, the classification network $f$ achieves an in-domain accuracy of $94.11\%$ so that we can call it a sketch expert. When evaluating the performance of the network on the real domain we see a drop to $9\%$ accuracy. For a 10-class problem, this performance is slightly below when predicting the classes uniformly at random.
This confirms that the domain gap between black and white sketch images and RGB photographs is notably bigger than the domain gap considered in\cite{mabu_semi-supervised_2021}.

\begin{figure}
    \centering
    \includegraphics[width=0.4\textwidth]{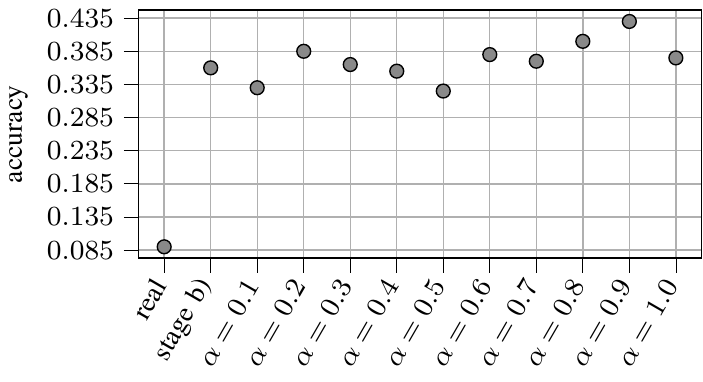}
    \caption{Classification results of the sketch expert network with different types of input images. The performance when an RGB photograph is given as input, is indicated by ``real''. CycleGAN-only performance is depicted with ``stage b)'' (our method when no task awareness is added). With increasing weighting factor $\alpha$ the generator which generates the inputs was trained with more emphasis on the task loss $\mathcal{L}_\mathit{task}$. The results are based on $70$ GT ($10\%$) images during stage c) training.}
    \label{fig:sketch_performance}
\end{figure}

Feeding $f$ with images generated by the generator $G_{\mathcal{R}\to\mathcal{S}}$ after stage b) training, already improves the accuracy substantially by $27$ percent points (pp) reaching an absolute accuracy of $36\%$.
When continuing training the generator with stage c), and feeding the images generated by that generator to the downstream task network $f$, we achieve up to $43\%$ accuracy depending on how much we weight the task loss component in \cref{eq:extendedloss}. A qualitative comparison of the network performance with respect to different inputs is shown in \cref{fig:sketch_performance}.
For values of $\alpha \leq 0.5$, we observe no clear trend compared to a CycleGAN-only training (only stage b), i.e.\ task agnostic). Whereas, we improve the accuracy using our method when $\mathcal{L}_\mathit{task}$ dominates ($\alpha > 0.5$) the adversarial loss, yielding a relative increase of up to $7$\,pp.

When we only use the task loss in the generator training  ($\alpha = 1.0$), the performance of $f$ drops again. This is expected as we remove the adversarial loss completely and therefore do not get notable feedback of the discriminator. We will also investigate the change of the performance with respect to $\alpha$ in our semantic segmentation experiments (see \cref{sec:semseg}) where we observe this trend more clearly in the CARLA setup (cf. \cref{fig:vary_alpha_175vs285_carla}).

\begin{figure}
    \centering
    \includegraphics[width=0.485\textwidth, trim=0 10 0 0, clip]{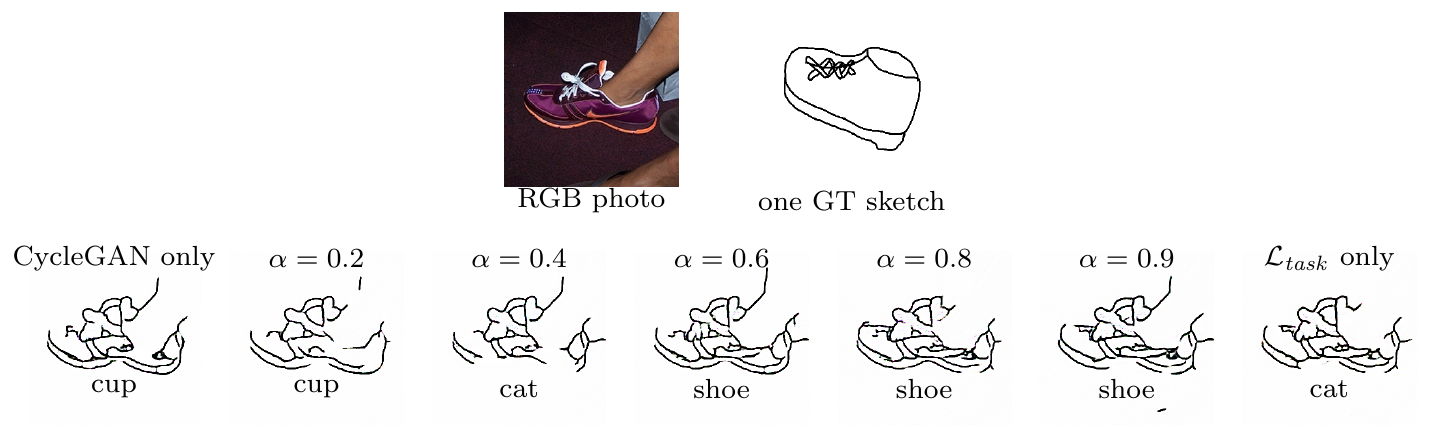}
    \caption{Quantitative results of the output of the generators trained with different weighting ($\alpha$) of the downstream task loss. Top row: Real domain RGB photo (input of generator) and one of the ground truth sketches of the RGB photo. Bottom row: Generated sketches and the prediction results of the downstream task network.}
    \label{fig:qualative_comp_sketches}
\end{figure}

Exemplary, we show the results of different generators trained with different $\alpha$ weighting in \cref{fig:qualative_comp_sketches}. The classification results of the network trained on sketches are reported underneath the images.
Even though $f$ classifies the shoe correctly for $\alpha = 0.6$, $\alpha = 0.8$ and $\alpha =0.9$, we as human can barely see a difference between the generated images. It seems the images generated by a generator trained with higher $\alpha$ values might capture more details.
Nevertheless, these results show that with emphasis on the task loss the generator learned to support the downstream task.

In the next paragraph we consider a semantic segmentation task  -- a distinct more challenging task -- on the potentially smaller domain gap between real and simulated street scenes.

\subsection{Semantic segmentation on urban and simulated street scenes} \label{sec:semseg}
\paragraph{Dataset}\label{sec:dataset}
For the semantic segmentation task we focus on urban street scenes. We use the established dataset Cityscapes\cite{cordts_cityscapes_2016} for the real domain. The dataset contains images which were taken in multiple cities and have a resolution of $2,\!048 \times 1,\!024$ pixels. For our experiments we use the $2,\!975$ images of the train split as well as the $500$ validation images where the fine annotations are publicly available.

For the synthetic domain we conduct our experiments on two different datasets.
In the first experiment we use one of the standard dataset in domain adaptation experiments: SYNTHIA-RAND-CITYSCAPES (Synthia)\cite{ros_synthia_2016}, a dataset based on the Unity simulation engine.
It consists of $9,\!000$ images with a resolution of $1,\!280 \times 760$, randomly taken in a virtual town from multiple view points.
To have coincided classes in both domains, we restrict the classes to the commonly used $16$ for domain adaptation which are the Cityscapes training IDs except for \textit{train}, \textit{truck} and \textit{terrain}\cite{brehm_semantically_2022}.
As no fixed validation set is given, we leave out the last $1400$ images during training. Using the first $700$ for validation.
This leads to camera view points for evaluation that are independent of those used for training.

As a second setup we generated a dataset with the help of the open-source simulator CARLA\cite{dosovitskiy_carla_2017} which bases on the Unreal Engine.
The simulator allows for the extraction of a strongly controlled dataset to realize our hypothesis of unlimited data in the synthetic domain.
To showcase this we use only one town map of CARLA (town 1) with fixed environmental settings like weather, wind etc.
We generated $3,\!900$ images for training and $1,\!200$ images for validation with a resolution of $1,\!920\times1,\!080$ by randomly spawning the ego vehicle on the map. Furthermore, we spawned a random number of road users for a diverse scenery. Similar to Synthia not all Cityscapes training classes exist in CARLA. Especially there is no distinction between different vehicle and pedestrian types. To this end, we fuse them into a vehicle and pedestrian metaclass. Therefore, we consider only $13$ classes: \textit{road}, \textit{sidewalk}, \textit{building}, \textit{wall}, \textit{fence}, \textit{pole}, \textit{traffic light}, \textit{traffic sign}, \textit{vegetation}, \textit{terrain}, \textit{sky}, \textit{pedestrian}, \textit{vehicles}.

In contrast to manually labeling, the segmentation mask of CARLA is very fine detailed. To adapt the more coarse labeling of a human annotator and therefore generate more comparable semantic segmentation masks we smooth the label and the RGB images in a post-processing step according to the method from\cite{rottmann_automated_2022}.

\paragraph{Implementation details \& results -- synthetic domain expert}
For the semantic segmentation network $f$, we use the PyTorch\cite{paszke_pytorch_2019} implementation\footnote{\centering \href{https://github.com/pytorch/vision/blob/main/torchvision/models/segmentation/deeplabv3.py}{https://github.com/pytorch/vision/blob/main/\-torchvision/models/\-segmentation/\-deeplabv3.py}} of a DeepLabv3 with ResNet101 backbone\cite{chen_rethinking_2017} ranging under the top third of semantic segmentation models on Cityscapes with respect to the comparison of Minaee et al.\cite{minaee_image_2021}.
We train the semantic segmentation network from scratch without pre-training to evaluate the domain gap accurately.
To range the results, we trained and evaluated $f$ once on Cityscapes to state the oracle performance of the from-scratch-trained network independently of our experiments. This led to a mIoU of $62.74\%$ on the Cityscapes validation set. This model is only used as reference and therefore we refrained from hyperparameter tuning.

For both experiments, we train using Adam\cite{kingma_adam_2014} with class weighting and normalize the images according to the pixel-wise mean and variance of the training data. During training, we used a horizontal flip with a probability of $50\%$ and a polynomial learning rate (LR) scheduler.
For the experiments with Synthia as synthetic domain, we trained $f$ for $3$ days with a batch size of $2$ due to GPU memory capacity which led to $107$ epochs of training on the training dataset. During training, we crop patches of size $1,\!024\times512$.

To measure the performance of our method we use the established mean intersection over Union (mIoU) metric which measures the relation between the intersection and the union of the prediction and ground truth as follows:
\begin{align*}
    \mathit{IoU}(c) &= \dfrac{\mathit{TP_c}}{\mathit{TP_c} + \mathit{FP_c} + \mathit{FN_c}}, \quad c = 1, \ldots, C \\
    \mathit{mIoU} &= \dfrac{1}{C}\sum_{c=1}^C \mathit{IoU}(c)
\end{align*}
where $\mathit{TP_c}$ denotes the amount of pixels correctly classified as class $c$, $\mathit{FP_c}$ the amount of pixels which were falsely predicted as class $c$ and $\mathit{FN_c}$ denotes the amount of omitted predictions, so pixels which are of class $c$ but were not classified as such.
On the in-domain validation set we achieve a mIoU of $64.83\%$.

For the experiments with CARLA as synthetic domain, we trained our network for $200$ epochs with random quadratic crops of size $512$. We yield the best mIoU of $91.89\%$ on the validation set. Benefiting of the simulation we constructed a meaningful in-domain expert.

\begin{figure}
    \centering
    \includegraphics[width=0.35\textwidth]{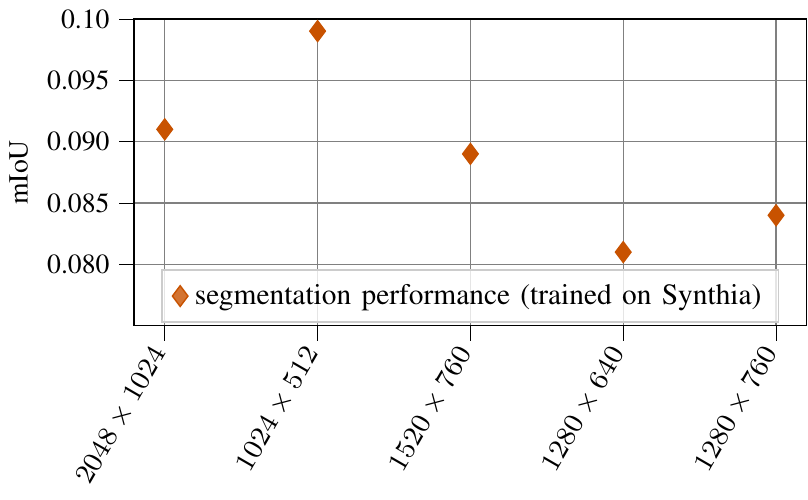}
    \caption{Prediction performance of the Synthia expert given differently scaled Cityscapes images as input. The resolution of the scaled images is given on the x-axis. Scaling needs to be considered as the aspect ratio and the resolution differ between training (synthetic domain) and inference on Cityscapes images (real domain).}
    \label{fig:scaling}
\end{figure}

As the image resolution of Synthia and CARLA images, differ from the resolution of Cityscapes, a resizing is necessary. Depending on the scaling and aspect ratio the network's prediction performance differs. We chose the scaling with the best performance, which we found for $1,\!024\times512$.
A comparison of the network's input resolution (x-axis) and network's performance measured by mIoU (y-axis) for the Synthia expert is shown in \cref{fig:scaling}.
For a fair comparison we let the GANs generate the same resolution.

\paragraph{Implementation details -- domain shift}
For the I2I (stage b)), we use the same architecture as described in the CycleGAN publication\cite{zhu_unpaired_2017}.
When not denoted otherwise we used $175$ epochs for the stage b) training (task agnostic training) and additionally $50$ epochs for the stage c) training where labeled data is available. We use the pixel-wise cross entropy as task loss.
To balance the scale of the task loss $\mathcal{L}_\mathit{task}$ with respect to the adversarial generator loss $\mathcal{L}_{G_{\mathcal{R}\to\mathcal{S}}}$ we include an additional scaling factor $\gamma$. Multiplying the task loss with $\gamma$ leads to more balanced loss components and therefore a better interpretability.

\paragraph{Experiment setup and results} \label{sec:semseg_results}
First experiments were done on a mixture of labeled and unlabeled data, but we experienced an unstable training when alternating between the corresponding loss functions $\mathcal{\Tilde{L}}_\text{Gen}$ and $\mathcal{L}_\text{Gen}$. Splitting the generator training into two stages as described in \cref{sec:method}, led to a more stable training and therefore better results.

\begin{figure}[bt]
    \centering
    \includegraphics[width=0.48\textwidth]{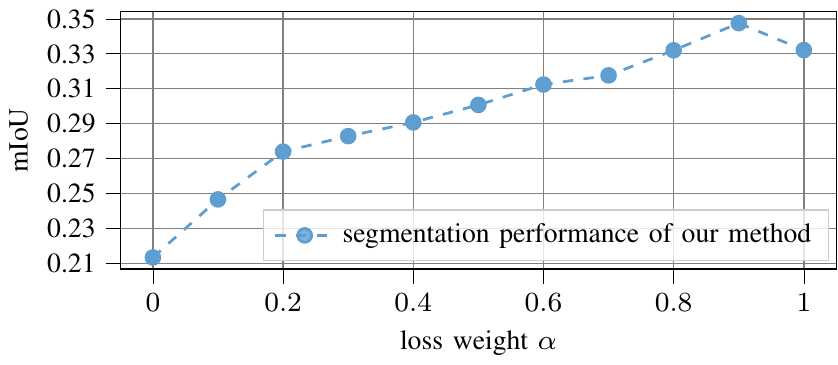}
    \caption{Influence of the task loss based on the Synthia experiment setup. The weighting factor $\alpha$ represents a linear interpolation between the adversarial generator loss and the task loss (cf.\!\cref{eq:extendedloss}), Resulting in the original CycleGAN implementation for $\alpha = 0$ and only the pixel-wise cross entropy loss for $\alpha=1$}
    \label{fig:alphaSynthia}
\end{figure}

\begin{figure*}[tb]
    \centering
    \includegraphics[width=\textwidth]{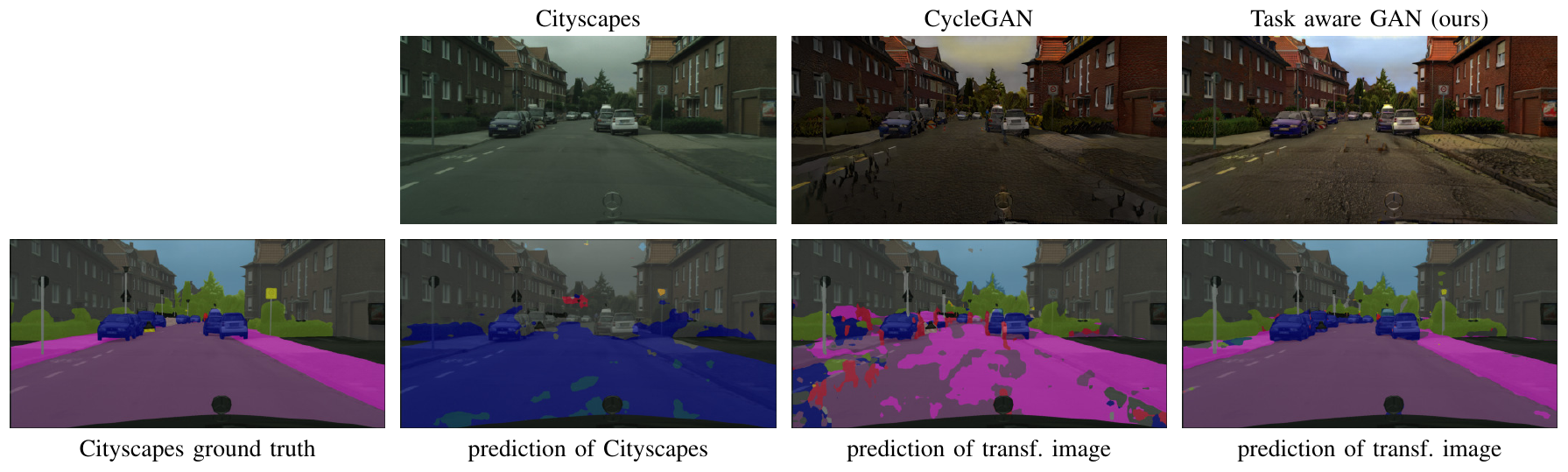}
    \caption{Comparison of prediction results of an untranslated Cityscapes image (left), task agnostic I2I (mid) and our approach (right) based on a semantic segmentation network trained on Synthia.}
    \label{fig:qualitative_res_synthia}
\end{figure*}

\begin{figure}[bt]
    \centering
    \includegraphics[width = 0.48\textwidth]{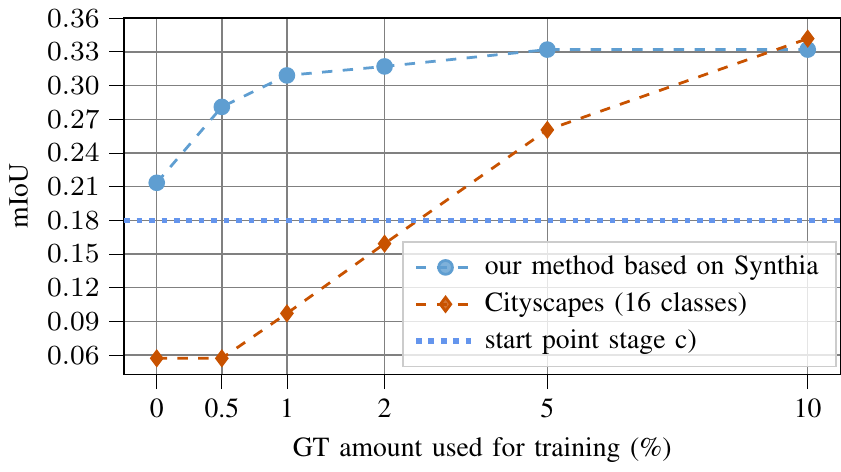}
    \caption{Performance comparison of our method based on Synthia setup with different amount of ground truth (blue) and a from scratch supervised training on Cityscapes with the same amount of data (orange). }
    \label{fig:data_vs_miou_synthia}
\end{figure}

\begin{figure}[bt]
    \centering
    \includegraphics[width=0.48\textwidth]{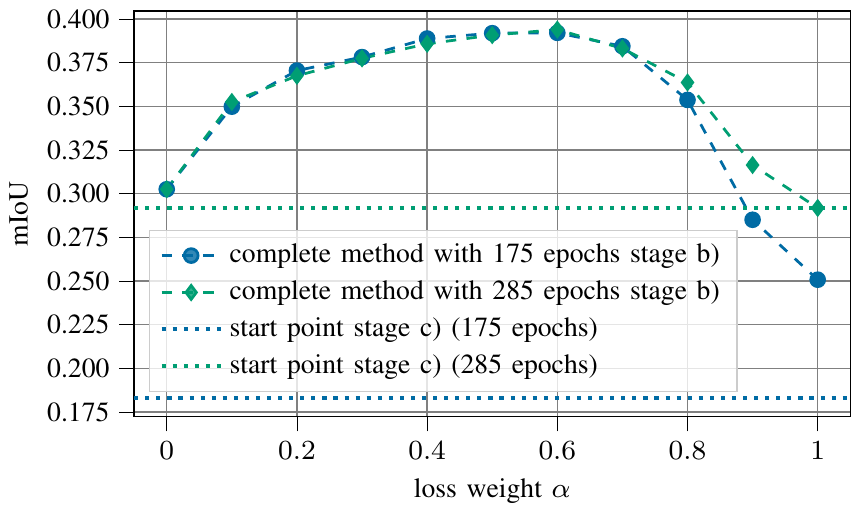}
    \caption{Influence of the task loss on the network performance for the CARLA experiment setup. On the x-axis the weighting factor ($\alpha$) of the task aware loss is plotted. The weighting represents a linear interpolation between the adversarial generator loss and the task loss \cref{eq:extendedloss}.
    Results after stage c) based on a 175 epochs unsupervised GAN-training are shown in blue.
    The green graph shows the method performance when trained with a longer amount of stage b) steps.}
    \label{fig:vary_alpha_175vs285_carla}
\end{figure}

\begin{figure}[bt]
    \centering
    \includegraphics[width = 0.48\textwidth]{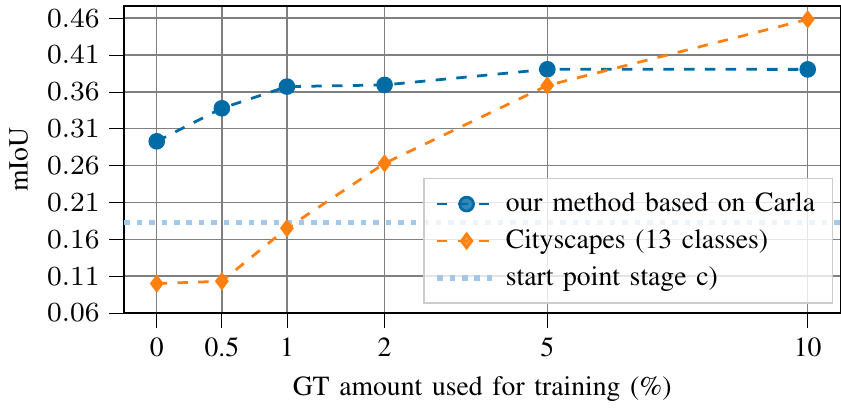}
    \caption{Performance comparison of our method based on CARLA setup with different amount of GT (blue) and $f_\mathcal{R}$ trained supervised from scratch on Cityscapes with the same amount of data (orange).}
    \label{fig:data_vs_miou_carla}
\end{figure}

\begin{figure*}[tb]
    \centering
    \includegraphics[width=\textwidth]{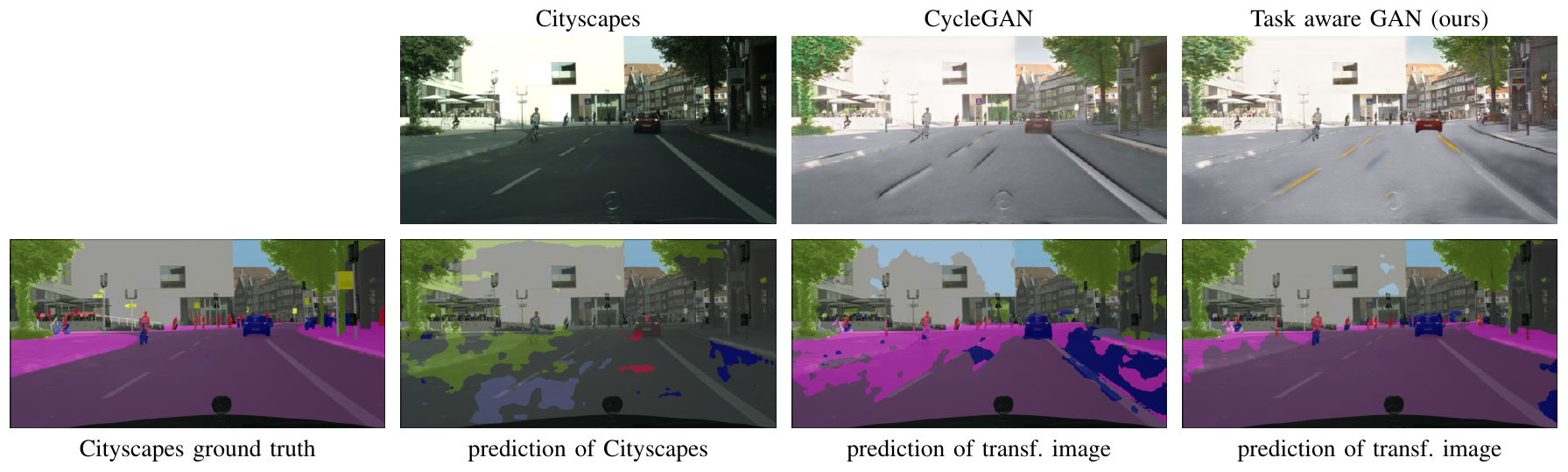}
    \caption{Comparison of prediction results of an untranslated Cityscapes image (left), task agnostic transfer (mid) and our approach (right) based on a semantic segmentation network trained on our CARLA dataset.}
    \label{fig:qualitative_res_carla}
\end{figure*}

To evaluate our approach we set up three experiments mainly differing in the stage c) training:
\begin{enumerate}
    \item Variation of $\mathcal{L}_\mathit{task}$ weighting (cf. \cref{fig:alphaSynthia,fig:vary_alpha_175vs285_carla})
    \item Evaluation of the synthetic domain expert on Cityscapes (cf. \cref{fig:qualitative_res_synthia,fig:qualitative_res_carla,tab:domain_gap})
    \item Variation of GT amount and comparison to the network performance when trained from scratch on Cityscapes with reduced GT (cf. \cref{fig:data_vs_miou_synthia,fig:data_vs_miou_carla})
\end{enumerate}
Within the experiments we compare our approach with the same types of methods as done in\cite{mabu_semi-supervised_2021}:
\begin{itemize}
    \item[M1] Synthetic domain expert  $f$ fed with images generated by a GAN with CycleGAN-only training (equaling $\alpha= 0.0$)
    \item[M2] Synthetic domain expert $f$ fed with real images without domain transformation (original Cityscapes images)
    \item [M3] Semantic segmentation network $f_{\mathcal{R}}$ trained from scratch in a supervised manner on the small amount of labeled real-world images available at stage c).
\end{itemize}
We report the best mIoU of $f$ evaluated on the output of $G_{\mathcal{R}\to\mathcal{S}}$ based on the Cityscapes validation set during GAN training.
To analyze the impact of the different loss components in the Synthia setup, we set the scaling parameter $\gamma$ empirically to $0.25$, we fix the GT amount to $5\%$ ($148$ labeled training images) and vary the weighting parameter $\alpha$ between $0$~and~$1$. As stated in \cref{eq:extendedloss} we use a linear interpolation between the losses. For the choice $\alpha=0$, the training loss equals the original CycleGAN loss. For the other extreme choice $\alpha=1$, the adversarial loss part is omitted entirely. The corresponding results are shown in \cref{fig:alphaSynthia}.

We see the positive impact of the task awareness in the growing mIoU values. Using a weighting of $\alpha =0.9$ for $\mathcal{L}_\mathit{task}$, we achieve $34.75\%$ mIoU which is a performance increase of $13.41$\,pp compared to M1 (task agnostic GAN training).
In \cref{fig:qualitative_res_synthia} we show for one image from the Cityscapes validation dataset the differently generated images as well as their predictions by $f$ trained on Synthia.
The column ``Cityscapes'' in \cref{fig:qualitative_res_synthia} gives an exemplary visualization of the low prediction performance of a synthetic domain expert when never having seen real-world images (M2). The network's performance drops to roughly $10\%$ when real world images are used as input for the domain expert.
Here we see a significant difference to results reported by other domain adaptation methods which use ImageNet pre-trained networks, e.g.,\cite{dundar_domain_2018}. The domain gap is summarized in \cref{tab:domain_gap} where we compare ImageNet pre-trained network performance to ours evaluated on Cityscapes.
We state out-of-domain performance (second column), i.e.\ trained on Synthia, oracle performance (third column), i.e.\ trained on the full Cityscapes training dataset, and the domain gap between them, measured as difference in performance (last column). The results indicate that the ImageNet pre-training already induces a bias towards the real domain distorting a pure domain separation which should be pursued when analyzing domain gaps.
\begin{table}[tb]
    \caption{Domain gap comparison of networks trained from scratch vs ImageNet pre-trained}
    \centering
    \begin{tabular}{c c c c}
        \hline
        \multicolumn{4}{c}{Synthia $\to$ Cityscapes (mIoU in \%) }\\
        \hline
         method & out-of-domain & oracle & gap\\
         \hline
        ImageNet pre-trained\cite{dundar_domain_2018} & 31.8 & 75.6 & 43.8\\
        from scratch (ours) &  9.9 & 62.7 & 52.8
    \end{tabular}
    \label{tab:domain_gap}
\end{table}
Based on our training-from-scratch setup, using task agnostic generated images (M1) improves already significantly the performance ($11.44$\,pp)  whereas our approach (task aware GAN training) can lead to a relative improvement of up to $24.85$\,pp when $5\%$ GT images are available.

Moreover, we analyze the capacity of the method based on the amount of GT available. Therefore, we fix $\alpha = 0.8$ and vary the GT amount for the stage c) training in our method. We randomly sample images from the Cityscapes training dataset for each percentage but fix the set of labeled data for the experiments with CARLA and ``Cityscapes-only'' training (M3) for the sake of comparison.
Results are shown in \cref{fig:data_vs_miou_synthia} (blue curve), the graphs depict the ground truth amount in comparison to the mIoU. The dotted horizontal line is the mIoU achieved by $f$ when exclusively feeding images generated by the task agnostic GAN after finishing stage b) training for $175$ epochs which we use as start point for stage c) training. For a fair comparison we trained the task agnostic GAN for another $125$ epochs which results in a better mIoU of $21.34\%$ in epoch $243$ which we use as result for GT $= 0$ (equaling $\alpha=0.0$). For our experiment we compare $0.5\%$ ($14$ images), $1\%$ ($29$ images), $2\%$ ($59$ images), $5\%$ ($148$ images) and $10\%$ ($297$ images) of GT data for the stage c) training.
Triggering the task awareness with only $14$ images already improves the network accuracy by $6.75$\,pp.
The graph monotonously increases with the amount of GT available, reaching $33.20\%$ mIoU for $5\%$ GT and $33.76\%$ when the GAN is trained with $10\%$ labeled data in stage c).
The results show that training $G_{\mathcal{R}\to\mathcal{S}}$ with the task loss on negligible few GT data, improves the neural network's understanding of the scene without retraining the neural network itself.

Additionally, we compare our method with results of $f_\mathcal{R}$ (M3). We train the semantic segmentation network from scratch using (only) the same images as for the stage c) training of our method.
Having no labeled data available, a supervised method can barely learn anything. Therefore, we set the value to the same as for $0.5\%$ GT which most likely overestimates the performance. The results are visualized by the orange curve in \cref{fig:data_vs_miou_synthia}.
The results show that our method outperforms M3 by a distinct margin when only a few labels are available.
However, when more than $297$ (10\%) fine labeled images of Cityscapes are available, a direct supervised training should be taken into consideration.

%%%%%%%%%% Carla %%%%%%%%%%%%%
For the CARLA experiments we set $\gamma=1$, as the losses are already in the same scale.
We repeat the experiments 1) - 3) on our CARLA dataset with the CARLA semantic segmentation expert. The results of varying $\alpha$ are visualized by the blue curve in \cref{fig:vary_alpha_175vs285_carla}.
Also, on the CARLA dataset our method shows a notable improvement over CycleGAN-only training (M1; $\alpha=0$) when choosing a balanced weighting between the adversarial and the task loss.
These experiments confirm that the task awareness improves the performance but should be used in addition and not as a stand-alone concept.

The results of experiment 2) and 3) are shown in \cref{fig:data_vs_miou_carla} where the blue curve represents the best mIoU results achieved with our method for the given GT amount and the orange curve depicts the results $f_\mathcal{R}$ (M3) given different amount of GT.
We see a slightly lower increase ($3.52$\,pp) of the performance between an exclusive stage b) training ($0\%$ GT, equaling a CycleGAN-only training) and our method with $0.5\%$ GT. The best mIoU is achieved when using $5\%$ of the dataset for the stage c) training, yielding a performance increase of $8.81$\,pp and an absolute mIoU of $39.06\%$.
For the direct supervised training we experienced a steeper increase when training only on the $13$ classes used for the CARLA experiments. However, As before our method outperforms M1 as well as M3 when less than 5\% GT is available. Above that, the supervised method is superior.

The qualitative results shown in \cref{fig:qualitative_res_carla} reveal a distinct improved semantic segmentation of the street scene (bottom row) on the CARLA dataset. As in the previous experiment visual differences recognizable by humans of the generated images with CycleGAN (top row mid) and our method (top row right) are limited. Furthermore, we see again the low performance caused by the domain gap when inputting real images to our from-scratch-trained synthetic expert $f$. On the untranslated images (M3), $f$ yields an mIoU of $9\%$.
Hence, the observed results achieved by CycleGAN (M1) and our method demonstrate a significant reduction of the domain gap.

Lastly we consider a longer stage b) training to find out whether our unsupervised pre-training was long enough or if a longer training further improves the results. We train in total $285$ epochs in stage b) and show the results of the complete method with $5\%$ GT in \cref{fig:vary_alpha_175vs285_carla} visualized by the green curve. The experiments reveal that the choice of longer unsupervised training is not game changing.
Although we start the stage c) training with a higher mIoU (dotted lines) when we train with stage b) for more steps, the experiments show that we achieve nearly the same absolute mIoU values, which only diverge when we put too much weight on the task loss ($\alpha \geq 0.8$). We therefore conclude that a moderate number of epochs is enough for stage b).

%%%%%%%%%%%%%%%%%%%%%%%%%%%%%%%%%%%%%%%%%%%%%%%%%%%%%%%%%%%%%%%%%
% DISCUSSION
%%%%%%%%%%%%%%%%%%%%%%%%%%%%%%%%%%%%%%%%%%%%%%%%%%%%%%%%%%%%%%%%%
\section{Conclusion and outlook} \label{sec:discussion}
In this paper, we presented a modular semi-supervised domain adaptation method based on CycleGAN where we guide the generator of the image-to-image approach towards downstream task awareness without retraining the downstream task network itself. In our experiments we showed on a ``real to sketch'' domain adaptation classification task that the method can cope with large domain gaps. Furthermore, we showed that our method can be applied to more complex downstream tasks like semantic segmentation yielding significant improvements compared to a pure image-to-image approach and from scratch training when a limited amount of GT is available.
In addition, we analyzed the impact of the task awareness and the amount of GT. We showed that with our approach, we significantly reduce the amount of manually labeled data needed on the real domain compared to a from scratch training.

Contrary to the common practice, all results were produced based on a non-biased domain gap. To this end, we trained all components from scratch.
Our achieved results suggest that the commonly used ImageNet pre-trained backbone already incorporates real world domain information and therefore distort the gap analysis.
Additionally, we showed that we can achieve very strong models when considering abstract representations (like sketches or modifiable simulations).

For future work, we are interested in elaborating more on the (intermediate) abstract representation, e.g., investigating if the robustness of the model can benefit therefrom and an in-depth analysis of how much abstract representations facilitate generalization. Recent work of Harary et al.\cite{harary_unsupervised_2022} strengthens our hypothesis of an enlarged generalizability.
Additionally, it has potential to help us better understand which visual features are important for a downstream task network. Generating more informative images for a downstream task network might give insights into the network behavior and help generate datasets which are cut down to the most important aspects of the scene for a neural network which is not necessarily what a human would describe as meaningful.
Moreover, an uncertainty based data selection strategy for stage c) training could further improve the method, e.g., active learning-based methods like Singh et al.\cite{singh_improving_2021} did for classification.
In addition, the method could be combined with self-training as these models need a good initialization to generate reasonable pseudo labels\cite{mei_instance_2020}. Nevertheless, when training the downstream task network completely from scratch, we have shown that the network performance is questionably low. Therefore, our method can be seen as complementary to the self-training approaches to ensure a reasonable prediction of the network in early self-training stages.

%%%%%%%%%%%%%%%%%%%%%%%%%%%%%%%%%%%%%%%%%%%%%%%%%%%%%%%%%%%%%%%%%
% ACK
%%%%%%%%%%%%%%%%%%%%%%%%%%%%%%%%%%%%%%%%%%%%%%%%%%%%%%%%%%%%%%%%%
\section*{Acknowledgment}
This work is funded by the German Federal Ministry for Economic Affairs and Climate Action. \\
The authors gratefully acknowledge the Gauss Center for Supercomputing e.V. (www.gauss-centre.eu) for funding this project by providing computing time through the John von Neumann Institute for Computing (NIC) on the GCS Supercomputer JUWELS\cite{julich_supercomputing_centre_juwels_2019} at Jülich Supercomputing Centre (JSC).
Furthermore, we thank Hannah Lörcks for assisting in setting up the classification experiments.

%%%%%%%% References %%%%%%%%%%%%%%
\bibliographystyle{IEEEtran}
\bibliography{lit_guidedGan_v4}

\end{document}